\documentclass[sigplan,10pt]{acmart}
\setcopyright{none}
\renewcommand\footnotetextcopyrightpermission[1]{} 
\usepackage{tikz}
\usepackage{amsmath}
\usepackage{subcaption}
\usepackage{listings}
\usepackage{graphics}

\usepackage{url}

\usepackage{breakurl}

\usepackage{multirow, makecell}

\newcommand{\remove}[1] {}

\newcommand{\fullpaper}[1] {}

\newcommand{\code}[1] {\texttt{#1}}

\makeatletter
\lst@Key{countblanklines}{true}[t]%
    {\lstKV@SetIf{#1}\lst@ifcountblanklines}
    
\lst@AddToHook{OnEmptyLine}{%
    \lst@ifnumberblanklines\else%
       \lst@ifcountblanklines\else%
         \advance\c@lstnumber-\@ne\relax%
       \fi%
    \fi}
\makeatother
  
\begin{document}

\title{Improving Inference Performance of Machine Learning with the Divide-and-Conquer Principle}

\author{Alex Kogan}
\affiliation{%
  \institution{Oracle Labs}
}
\email{alex.kogan@oracle.com}

\begin{abstract}
Many popular machine learning models scale poorly when deployed on CPUs.
In this paper we explore the reasons why and propose a simple, yet effective approach
based on the well-known Divide-and-Conquer Principle to tackle this problem of great practical importance.
Given an inference job, instead of using all available computing resources (i.e., CPU cores)
for running it, the idea is to break the job into independent parts that can be executed in parallel,
each with the number of cores according to its expected computational cost.
We implement this idea in the popular OnnxRuntime framework and evaluate its effectiveness 
with several use cases, including the well-known models for optical character recognition (PaddleOCR)
and natural language processing (BERT).

\end{abstract}

\maketitle
\pagestyle{plain}

\section{Introduction}
We live in the era of unprecedented attention to machine learning (ML) from researchers and practitioners alike.
New ML models across a variety of domains (or modalities, such as video, images and text) 
are proposed nearly daily, the models grow bigger and more sophisticated, and their 
components are continuously revised to achieve better accuracy scores on various tasks.
While lots of attention is given to training efficiency and prediction accuracy, seemingly less effort is focused on 
making sure those models perform well when deployed in practice, i.e., during inference~\cite{DK21}.
As we demonstrate in this paper, some models scale poorly (and at times, even worse!) when 
the number of available cores in a CPU-based deployment is increased.

Why does not the inference on CPUs scale? 
There are a variety of reasons, and we devote the entire section of this paper to look into some of them.
Briefly, they range from the micro-level, such as the use of
non-scalable operators inside ML architectures, to macro-level, such as employing ML architectures
that process input iteratively.

To mitigate those scalability challenges, one might consider redesigning their ML architecture or reimplementing
its non-scalable operations with a more efficient version. 
Such approaches, however, require either substantial ML domain specific expertise, exceptional engineering skills
and familiarity with ML frameworks used for inference,
significant investments (e.g., to retrain a new model, with a potential risk to the accuracy metrics), or all of the above.

In this paper, we take a different approach and propose to leverage the poor scalability of ML models by 
applying the Divide-and-Conquer Principle, a well-known algorithm design technique in Computer Science~\cite{CLR09}.
Specifically, instead of allocating all available computing resources (CPU cores) to the entire problem,
we propose to divide the problem into smaller chunks\footnote{We note that
unlike the classical Divide-and-Conquer Principle~\cite{CLR09}, we divide the problem only once, although 
it might be possible in some cases to divide it recursively into increasingly smaller chunks that can be executed by one thread each.}, 
let the framework decide how the computing resources should be
allocated among those chunks and then run their respective computations in parallel.
We argue that in many use cases, such a division is natural and requires only trivial changes in the user code.
We also describe a simple mechanism that allocates computing resources based on the expected
computational intensity (or weight) of each chunk.

Consider, for instance, a model for solving a natural language processing (NLP) task such as tweet classification.
Our approach allows efficient batching of inference requests of various sizes, eliminating the need for padding 
(a common, but wasteful solution to deal with batches of requests of variable size)
and letting the framework allocate computing resources proportionally to the length of each sequence.
We implement the aforementioned allocation mechanism in OnnxRuntime~\cite{OnnxRuntime}, a popular framework
for training and inferencing ML models, and extend its inference API to allow user code to invoke 
parallel inference on multiple inputs.
We demonstrate the effectiveness of our approach with several use cases, including highly popular models for image processing
(PaddleOCR~\cite{DLG20}) and NLP tasks (BERT~\cite{DCL19}).

The remainder of this paper is organized as following.
In Section~\ref{sec:why-slow} we elaborate on various reasons for why the inference (on CPUs) commonly does not scale well.
Next, we describe in Section~\ref{sec:dacp} the concept and implementation details of the Divide-and-Conquer Principle as it applies to inference.
Following that, we present in Section~\ref{sec:use-cases} several use cases of ML models where this principle can be applied, 
along with the performance evaluation of its benefits.
We discuss related work in Section~\ref{sec:related} and conclude the paper in Section~\ref{sec:discussion}.

\section{Why is Inference Slow?}
\label{sec:why-slow}
There are numerous reasons for this lack of scalability.
In this section we survey some of them.

\subsection{Not ``enough'' work}
One reason is simply because the amount of computation required by a model during inference is not ``enough'' 
for efficient parallelization.
As noted by Aminabadi et al.~\cite{ARZ22}, kernel implementations of various ML operations are often geared towards training,
which tends to consist of sizable batches of large inputs (e.g., sentences of 512 tokens).
During inference, however, the batches tend to be much smaller, and often include just one input (e.g., for real-time / interactive 
inference).
Besides, the inputs themselves can be small, e.g., a tweet or chatbot interaction consisting of just a few words.

Consider, for instance, highly popular Transformer-based~\cite{VSP17} models for NLP tasks, 
such as BERT~\cite{DCL19} or GPT-3~\cite{BMR20}, which rely mostly (but not solely) on
matrix multiplication primitives.
Those primitives are known to scale well for large matrices~\cite{DK21, MVG07, WWW21}.
However, when the actual input to the model during inference is short, 
matrix multiplications involve smaller and therefore, less amendable to efficient parallelization, matrices~\cite{DK21, MAH16, WWW21}.

\subsection{Non-Scalable Operators}
Another reason for poor scalability of some ML models is the use of non-scalable (and often, sequential) operators in their architecture.
Typically, the overhead of those operators would be negligible compared to other, more scalable parts of the model.
Yet, as the number of cores increases and following directly from the Amdahl's Law~\cite{Amd67}, their negative impact 
of non-scalable operators on the overall inference performance would grow.
Considering again the Transformer-based~\cite{VSP17} models mentioned above, Dice and Kogan have observed that
while matrix multiplication scales well, at least for long inputs, other operations such as layer normalization and softmax 
do not, contributing to the overall poor scalability of those models~\cite{DK21}.
In this paper, we consider a vision-based model, which employs sequentially implemented functions for 
internal format conversions, which similarly cause the entire model not to scale.

We note that some of those cases could be considered a performance bug in the underlying ML framework, 
which could be fixed by reimplementing the respective operators with more efficient (and parallel) alternatives.
This, however, requires lots of engineering effort, which includes performance analysis and deep understanding of 
corresponding framework implementation details.
Besides, some of the ML operators, such as layer normalization~\cite{BKH16}, require careful coordination among computing threads (e.g., to compute 
variance and standard deviation of all the hidden units in a layer and then use those statistics to normalize the values of the units)
and therefore do not lend themselves naturally for efficient parallelization.

\subsection{Framework Overhead}
Somewhat related to the prior point, an ML framework might add small but measurable overhead in invoking
model operations. Most popular ML frameworks, such as PyTorch, Tensorflow or OnnxRuntime, support multiple backends 
for executing ML operations, targeting different hardware architectures (CPU, GPU, TPU), utilizing different BLAS libraries 
(MKL, OpenBLAS, oneDNN, etc.), different threading infrastructure (Intel TBB, pthreads, custom implementation, etc.), etc.
Dispatching appropriate kernel (implementation) for every operator is efficient, but is sequential and 
requires non-trivial amount of work, especially when the model is executed \emph{interactively}~\cite{DK21} 
(the default execution mode in PyTorch). 
This overhead becomes substantial as the actual execution time of the kernels reduces with the increased number of cores.

In addition to the above, various kernels might require specific memory layout for its input parameters (tensors), and the framework
would add appropriate dummy operators for input/output conversion or data preparation~\cite{WWW21}.
As we demonstrate later in this paper, these operators might add substantial overhead as well.

\subsection{Model Architecture}
Quite often the high-level architecture of an ML model itself plays a substantial role in causing inference not to scale.
For instance, some ML models, especially ones built for video and image processing (e.g.,~\cite{WS15, DLG20, MHL21}), are composed 
as a multi-phase pipeline. The first phase of the pipeline would typically identify the points of interest in the input (e.g.,
text boxes in an image or a moving object in a video), while subsequent phases would process those points 
(iteratively or as a batch) to solve the predefined problem (e.g., identify text in the boxes or 
classify the moving object in the video).
The inference latency of such models might grow linearly with the number of objects identified in the first phase.
Furthermore, if even one phase of the pipeline does not scale well, the scalability of the entire pipeline is impaired.

\subsection{Padding}
Batching multiple inputs and processing them at once is a well-known way of improving inference throughput~\cite{ARZ22, ZWZ22, FYZ21, APY20, CWZ17}.
In fact, multiple serving system for machine learning models (such as TensorFlow Serving~\cite{TFServing} or TorchServe~\cite{TorchServe}) 
include tunable parameters that configure how long an inference server can wait in order to batch as many input requests as possible. 
However, when inputs in a batch do not have exactly the same shape, they need to be padded to be processed efficiently,
since underlying kernels typically anticipate batches of homogeneous inputs.
The padding leads to reduced computational efficiency, since it is treated by kernels as the rest of the input,
even though the corresponding output produced by the model is dismissed.

\section{Divide-and-Conquer Principle Applied to Inference}
\label{sec:dacp}

In this section, we describe the application of the Divide-and-Conquer Principle~\cite{CLR09} to 
the inference of ML models at the conceptual level and as a concrete realization by implementing
it in the OnnxRuntime framework.
We note that applying this principle does not directly address the reasons for poor scalability detailed 
in the previous section.
In fact, the advantage of our approach is that one does not have to identify and/or fix any scalability 
bottlenecks in their models to rip the benefits of its underlying idea.

\subsection{Concept}
The basic idea is pretty straightforward --- consider a computation job $J$, which can be broken 
into $k$ independent parts, $j_1$, $j_2$, ..., $j_k$, which can be executed in parallel.
Assume we have an oracle assigning relative weight $w_i \in (0,1]$ corresponding to, e.g., 
the number of required floating point operations (FLOPs)
or single-thread latency of the computation job part $j_i$.
Finally, assume we have $C$ computing cores available.
We strive to allocate to each part the number of cores relative to its weight, namely, we assign 
$c_i = max\{1, \lfloor w_i * C \rfloor \}$ cores for the part $j_i$.
This effectively means allocating $c_i$ worker threads for $j_i$ since we later create one worker 
thread per core (as common in ML frameworks, including in OnnxRuntime).

Note that $\sum_{i=1}^k{c_i}$ might be larger than C. 
This is obvious when the number of job parts, $k$, is larger than C, but it is possible even when $k \leq C$.
This does not create a problem other than implying that some job parts will be run after other job parts have finished 
(rather than running them all in parallel).
At the same time, due to the rounding-down (floor) function intended to reduce the above possibility of oversubscription,
some unallocated cores might remain.
To avoid this waste of available resources, we sort all the job parts by their remaining unallocated weight, i.e.,
by $ w_i * C - \lfloor w_i * C \rfloor$, and assign one core to each part in the descending order, up until all cores 
are allocated. 
The C++-like pseudo-code for the entire algorithm is given in Listing~\ref{alg:allocation}.

\begin{figure}[t!]
\begin{lstlisting}[language=C++, caption=Thread allocation algorithm, label=alg:allocation, escapechar=|,deletendkeywords={next}, commentstyle=\color{blue}]
vector<int> allocate(vector<Tensor> inputs, int numCores) {
  vector<int> threadAllocation;
  vector<tuple<int, float>> threadUnallocatedWeight;

  int numInputs = inputs.size();
  int allocatedCores = 0;
  int index = 0; 
	
  int totalSize = 0;
  for (auto j_i : inputs) totalSize += j_i.size()
  
  for (auto j_i : inputs) {
    int numThreadsToUse = 1;

    if (numInputs <= numCores) {
      int size = j_i.size();
      float w_i = ((float)size) / totalSize;
      numThreadsToUse = floor(w_i * numCores);
    
      // this may happen due to flooring
      if (numThreadsToUse < 1) numThreadsToUse = 1;

      unallocatedWeight.add(
         make_tuple(index, w_i * numCores - numThreadsToUse));
    }                    
    threadAllocation.add(numThreadsToUse);

    allocatedCores += numThreadsToUse;
    index++;
  }

  if (allocatedCores < numCores) {
    // sort the vector in decreasing order by
    // comparing the second field in each tuple
    sort(unallocatedWeight, bySecondField);

    int nextToAdjust = 0;
    while (allocatedCores < numCores) {
      // fetch the first field in the `nextToAdjust` tuple
      index = 
         unallocatedWeight[nextToAdjust % numInputs].get(0);
      threadAllocation[index]++;
      allocatedCores++;
      nextToAdjust++;
    }
  }
  return threadAllocation;
}
\end{lstlisting}
\end{figure}

Naturally, the idea described above raises the question of how to assign relative weight to a job part $j_i$.
In all our cases considered in Section~\ref{sec:use-cases}, the weight is simply 
set proportionally to the size of input tensors.
Specifically, let $s_i$ be the size of the input tensor for job part $j_i$.
We set $w_i$ to $\frac{s_i}{\sum_{i=1}^k{s_i}}$,
essentially assuming that the amount of computation (expressed as the number of required FLOPs)
grows roughly linearly with the input tensors' size.
In general, however, assigning weight can be done with the help of a profiling phase and a lightweight classification 
mechanism, which associates job parts of the same (or similar) shape 
(as the one encountered during the profiling phase) to the relative
weight obtained during profiling.

\begin{figure*}[t!]
\begin{minipage}[t]{.45\textwidth}
\begin{lstlisting}[language=Python, caption=Original (shortened and edited for clarity) \code{TextRecognizer} class implementation from PaddleOCR, label=fig:PaddleOCR-orig, escapechar=|,deletendkeywords={next}, commentstyle=\color{blue}]
class TextRecognizer(object):
   def __init__(self, args):
      ...
      self.predictor = ort.InferenceSession(args.file_path)
      self.postprocess_op = build_post_process(args)
      ...

   def __call__(self, img_list):
      img_num = len(img_list)
      for beg_img_no in range(0, img_num, batch_num):
         end_img_no = min(img_num, beg_img_no + batch_num)
           
         inputs = prepare(img_list, beg_img_no, end_img_no)
           
         outputs = self.predictor.run(inputs)
           
         preds = outputs[0]
         rec_result = self.postprocess_op(preds)
         all_results.add(rec_result)
        
      return all_results
           
\end{lstlisting}
\end{minipage}%
\hspace{0.8cm}
\begin{minipage}[t]{.5\textwidth}
\begin{lstlisting}[language=Python, caption=Modified \code{TextRecognizer} class implementation (uses \code{prun}). Added or modified lines are in \color{red}{red}, label=fig:PaddleOCR-changes, escapechar=|,deletendkeywords={next}, commentstyle=\color{blue}]
class TextRecognizer(object):
   def __init__(self, args):
      ...
      self.predictor = ort.InferenceSession(args.file_path)
      self.postprocess_op = build_post_process(args)
      ...

   def __call__(self, img_list):
      img_num = len(img_list)
      for beg_img_no in range(0, img_num, batch_num):
         end_img_no = min(img_num, beg_img_no + batch_num)
           
         inputs = prepare(img_list, beg_img_no, end_img_no)
         |\color{red}all\_inputs.append(inputs)|
      |\color{red}all\_outputs = self.predictor.prun(all\_inputs)|
      |\color{red}for outputs in all\_outputs:|
         preds = outputs[0]
         rec_result = self.postprocess_op(preds)
         all_results.add(rec_result)
        
      return all_results
\end{lstlisting}
\end{minipage}
\end{figure*}

\remove{
Given that modern processors feature multiple virtual threads sharing one physical core 
(aka hyper-threads), the second question a reader may wonder about is how to set
$C$, the number of computing cores. The answer to this question depends on the
nature of the anticipated workload. If the workload is compute-intensive, e.g., 
consisting mostly of matrix multiplication operations, there is little to no benefit
from using more than one hyper-thread on any particular physical core as those
threads share the same pipeline (and thus the same arithmetic units).
Otherwise, if the workload is memory-intensive, there might be some benefit from 
overlapped work.
As our evaluation shows, in our use cases the use of hyper-threading did not change
the results substantially.}

\subsection{Implementation Details}
We extend the API of the \code{InferenceSession} class of OnnxRuntime with a new \code{prun} method.
This method is modeled after the existing \code{run} method 
used as the main entry point when running inference.
The main difference is that \code{prun} accepts a list of inputs (instead of just one) and 
returns a list of outputs.

Internally, the implementation of \code{prun} iterates over the list of inputs, calculates their
size (after validating those are tensors) and corresponding relative weight, and applies the 
allocation algorithm described in Listing~\ref{alg:allocation} to associate the number of 
worker threads with each input (job part). Following that, the implementation creates
one worker thread for each input, and runs them in parallel. 
Each worker thread, in turn, creates a thread pool of the size calculated by the allocation algorithm 
(the thread pool includes the worker thread itself), and invokes the \code{run} method of
the \code{InferenceSession} object with that thread pool.
The entire patch of the OnnxRuntime codebase to implement the \code{prun} functionality and
other minor internal changes (such as having the \code{run} method to accept a thread pool as 
an optional argument instead of always using the default pool) consisted of around 200 lines of code.

On the user side, the code also has to change to make use of the new \code{prun} API. 
Those changes, however, are quite straightforward. 
Instead of invoking \code{run} for every job, a user needs to create a list of job
parts and call \code{prun}. 
In addition, the user needs to rearrange the post-processing code to iterate 
over the results of \code{prun}, and apply any post-processing to each returned output (object).
As an example of what the user code changes entail, 
we show the original Python code (edited for brevity and clarity) of the TextRecognizer class in PaddleOCR
(Listing~\ref{fig:PaddleOCR-orig}) 
alongside the modified version that makes use of the new \code{prun} API (Listing~\ref{fig:PaddleOCR-changes}).

\section{Use Cases}
\label{sec:use-cases}

Before we detail the use cases where the Divide-and-Conquer Principle is beneficial
and report on our performance findings, we give a brief summary of our evaluation setup and methodology.
We run all our experiments on a 16-core AMD-based VM in Oracle Cloud (aka OCI VM.Standard.E3.Flex).
(We also ran some experiments on a newer E4 shape, but have not noticed substantial differences).
To reduce performance variability, especially as we create separate thread pools for the variants that
use \code{prun}, we use thread binding (pinning), for all the evaluated variants.
Every experiment was repeated $5$ times, and we report the mean.
We note that the standard deviation of all reported results, except for one specific case discussed below, was
extremely low (typically, less than 1\% of the mean).
For our experiments, we use the latest release versions (as of the date of writing this paper) of the corresponding software,
specifically OnnxRuntime v1.11.1 and PaddleOCR v2.5.

\subsection{Sequential Pipeline}
\label{subsec:paddle}

\begin{figure*}[t!]
\includegraphics[width=1\linewidth]{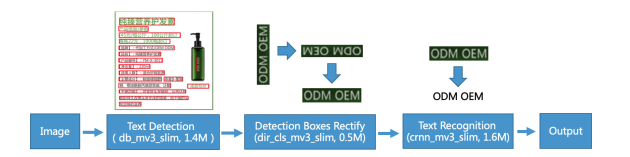}
\caption{PaddleOCR 3-phase pipeline (edited version of Figure~2 from~\cite{DLG20}).}
\label{fig:paddleOCR-pipeline}
\end{figure*}

Our first example of where applying the Divide-and-Conquer Principle is extremely useful is PaddleOCR~\cite{DLG20}.
PaddleOCR is a lightweight OCR system, which consists of three parts: Text Detection, 
Text Classification (called Detection Boxes Rectify in~\cite{DLG20}) and Text Recognition.
Each of those parts corresponds to a separate ML model.

The OCR pipeline accepts an image file and passes it first through the text detection phase whose objective is
to locate text areas in the image.
The output of this phase is a list of potential text boxes' coordinates.
Next, the list is iterated over, and each item in that list (i.e., a text box) is sent to the text classification model, which 
decides whether the box needs to be transformed into a horizontal rectangle box before the actual text recognition takes place.
Based on the classifier's decision, each box is altered respectively.
Finally, the list is iterated over again, and each item is sent to the text recognition model for inference, which
recognizes the text in the given box and produces the actual character sequence based on the supplied character dictionary.
This process is depicted in Figure~\ref{fig:paddleOCR-pipeline}, which is a redacted version of Figure~2 from~\cite{DLG20}.

\begin{figure}[t!]
\includegraphics[width=1\linewidth]{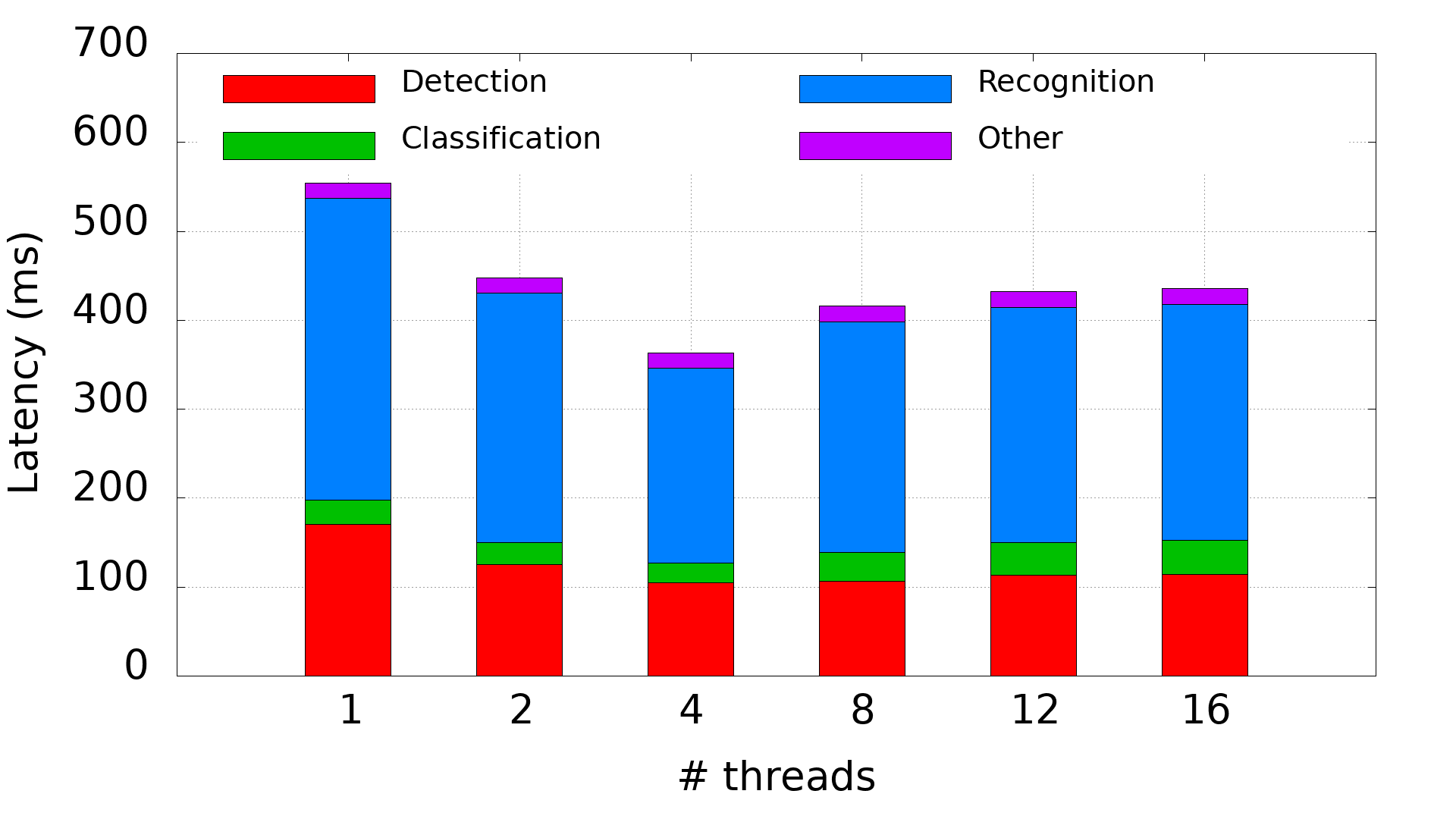}
\caption{Inference latency of PaddleOCR with a varying number of threads, broken down by the three phases of the pipeline.}
\label{fig:paddleOCR-base}
\end{figure}

In our experiments with PaddleOCR, we observe that the system does not scale well with 
the increase in the number of available cores.
We demonstrate that in Figure~\ref{fig:paddleOCR-base} depicting inference latency as a function of available cores 
(which directly translates into the number of worker threads used by the runtime).
For all experiments in this section, including the one in Figure~\ref{fig:paddleOCR-base},  we use
a subset of images from the OpenImages dataset~\cite{OpenImages}, selected according to a criterium described below.

In Figure~\ref{fig:paddleOCR-base}, we break the total latency into time spans corresponding to the three phases
of the OCR pipeline discussed above.
As one can notice, the average inference latency goes down from $554$ ms for $1$ thread 
to $364$ ms for 4 cores and then back up to $435$ ms for $16$ cores.
Interestingly, the Text Classification phase shows negative scalability, where it takes $27$ ms
to process an image, on average, with $1$ thread, but it takes $38$ ms to do the same with $16$ threads --- 
a slowdown of $1.4$x.
This shows an example of a system where, beyond a certain point, adding more threads 
not only does not help, but actually harms performance.
Discussing concrete reasons for the lack of scalability of these specific models is
not in the scope of this paper.
For a curious reader, however, we note that a built-in OnnxRuntime profiling tool shows inflated execution times for the 
output reordering operators (which are inserted by the framework, along with the input reordering operator, to 
convert the memory layouts of input arguments for various kernels).

We apply the Divide-and-Conquer Principle to the last two phases of the OCR pipeline, namely the 
Text Classification and Recognition. To that end, instead of invoking the corresponding models for each
text box produced by Text Detection, we send all the boxes to the runtime 
(by invoking the \code{prun} API) and effectively let the runtime decide how many cores / worker 
threads to allocate each box based on its relative size.
The required changes to implement this functionality in the Text Recognition phase are depicted in Listing~\ref{fig:PaddleOCR-changes};
the changes to the Text Classification phase are similar.

For our performance evaluation, we compare the \code{prun} implementation as discussed in Section~\ref{sec:dacp} (and
depicted in Listing~\ref{alg:allocation}), which we denote as \code{prun-def} on the charts, to a few simple variants.
The first variant, denoted as \code{prun-1}, simply allocates one worker thread to each input in the list given to \code{prun}.
The second variant, denoted as \code{prun-eq}, allocates an equal number of cores for each input (but at least one), i.e.,
sets $c_i = max\{1, \lfloor k / C \rfloor \}$.
Our motivation is to show that trivial solutions might also be useful in certain scenarios (as discussed below), yet they tend to
underperform compared to \code{prun-def}.

\begin{figure}[t!]
\includegraphics[width=1\linewidth]{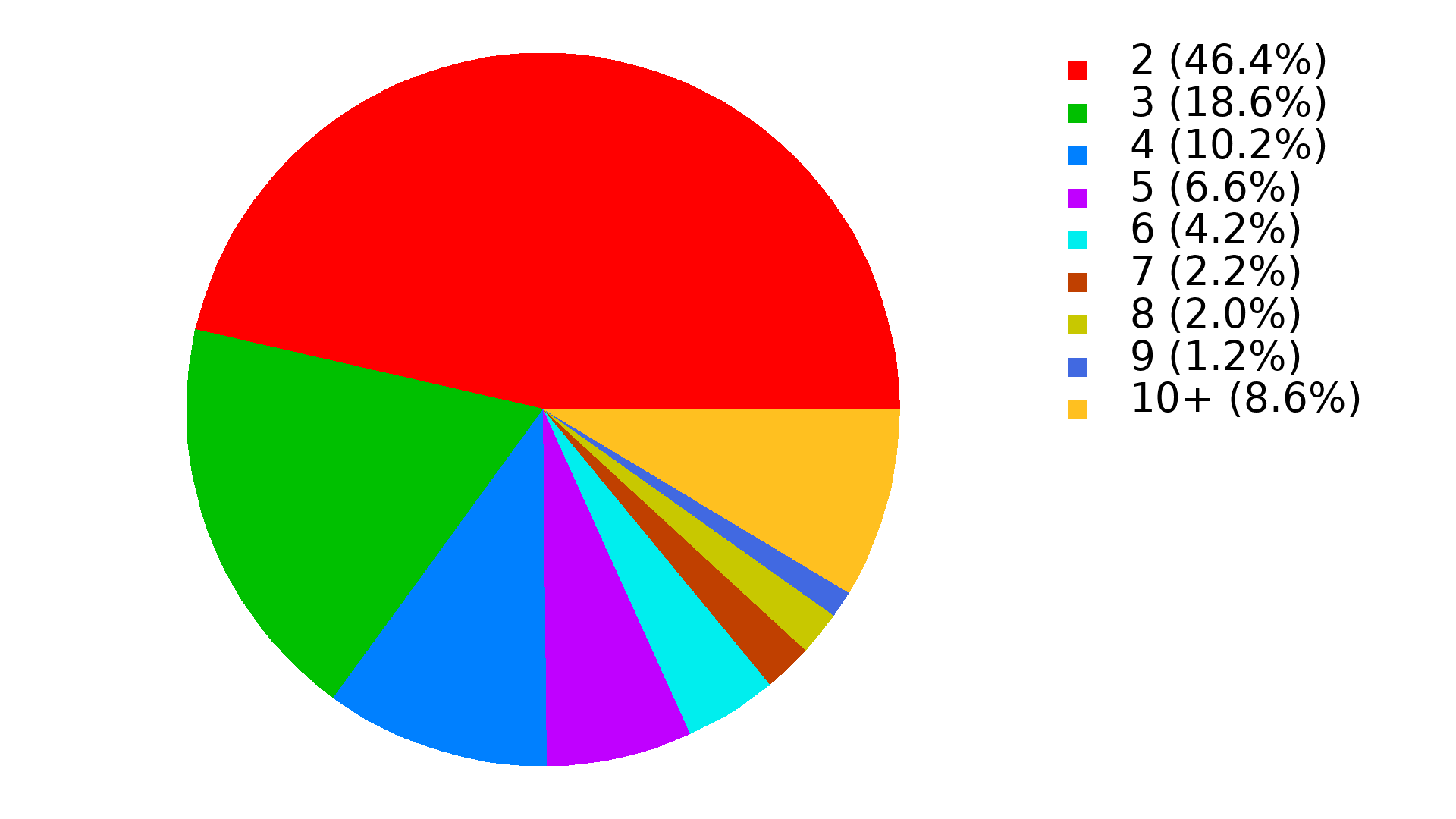}
\caption{Distribution of the number of detected text boxes in the input dataset.}
\label{fig:paddleOCR-piechart}
\end{figure}

We note that the benefit of \code{prun} in this use case is possible only when
there are at least two text boxes identified in the Text Detection phase. 
Otherwise, the other two phases would not be used (if no text boxes detected) or the \code{prun-def} variant 
will use the same (maximum) number of cores as the base (unmodified) version (if only one text box is detected).
As a result, the subset of images used for performance evaluation in this section includes images with at least two identified text boxes.
The pie chart in Figure~\ref{fig:paddleOCR-piechart} shows the distribution of the actual number of boxes detected in 
the first phase of the OCR pipeline for the entire dataset.
The total number of images in the dataset was $500$  -- this number was chosen 
to keep the evaluation times reasonably short.
(We note that we also ran evaluations on a larger dataset that includes images with less than two text boxes and
confirmed that the use of \code{prun} does not create any overhead in those cases.)

\begin{figure*}[t!]
\subfloat[][Text Classification]{\includegraphics[width=0.33\linewidth]{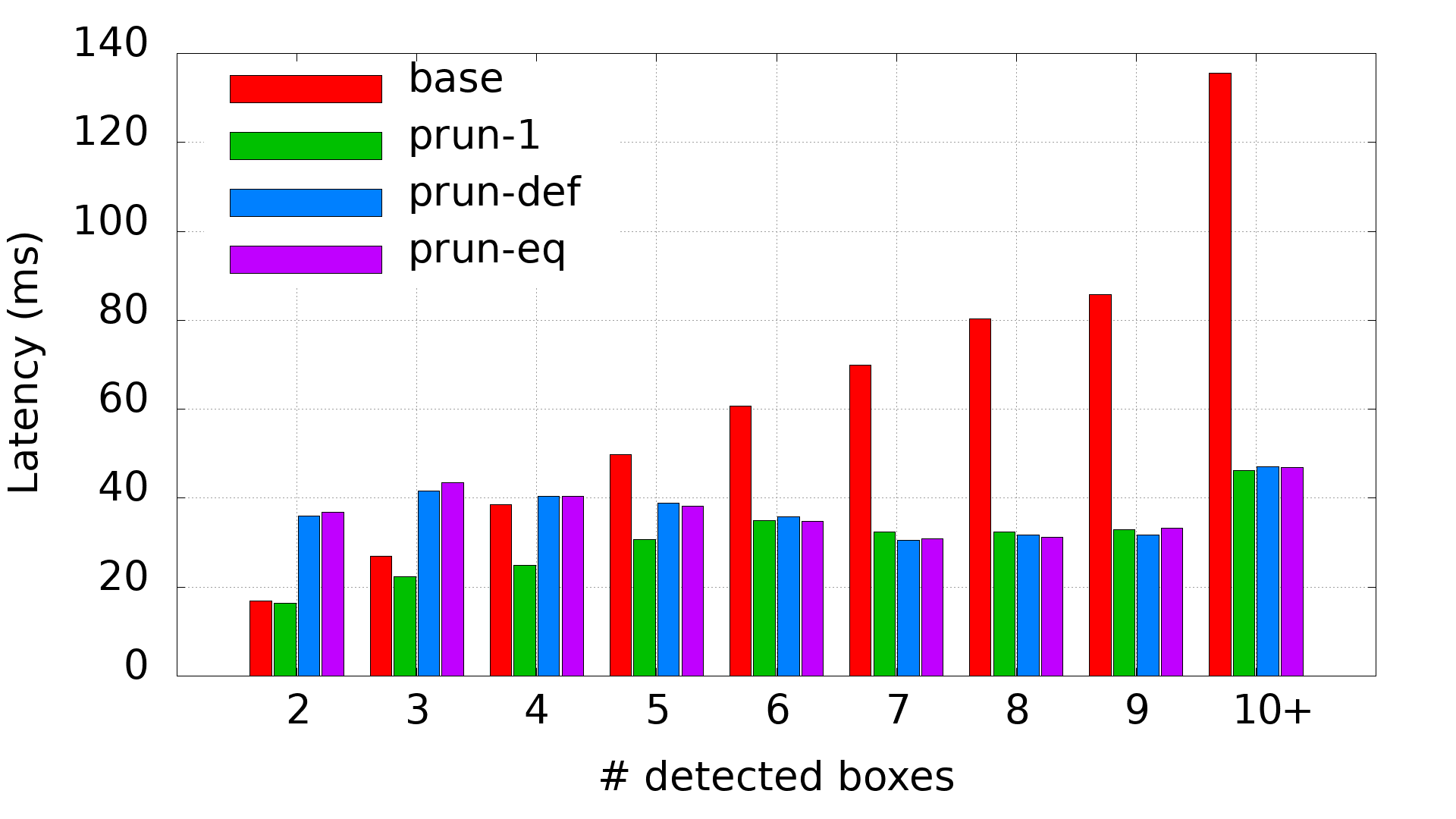}}
\subfloat[][Text Recognition]{\includegraphics[width=0.33\linewidth]{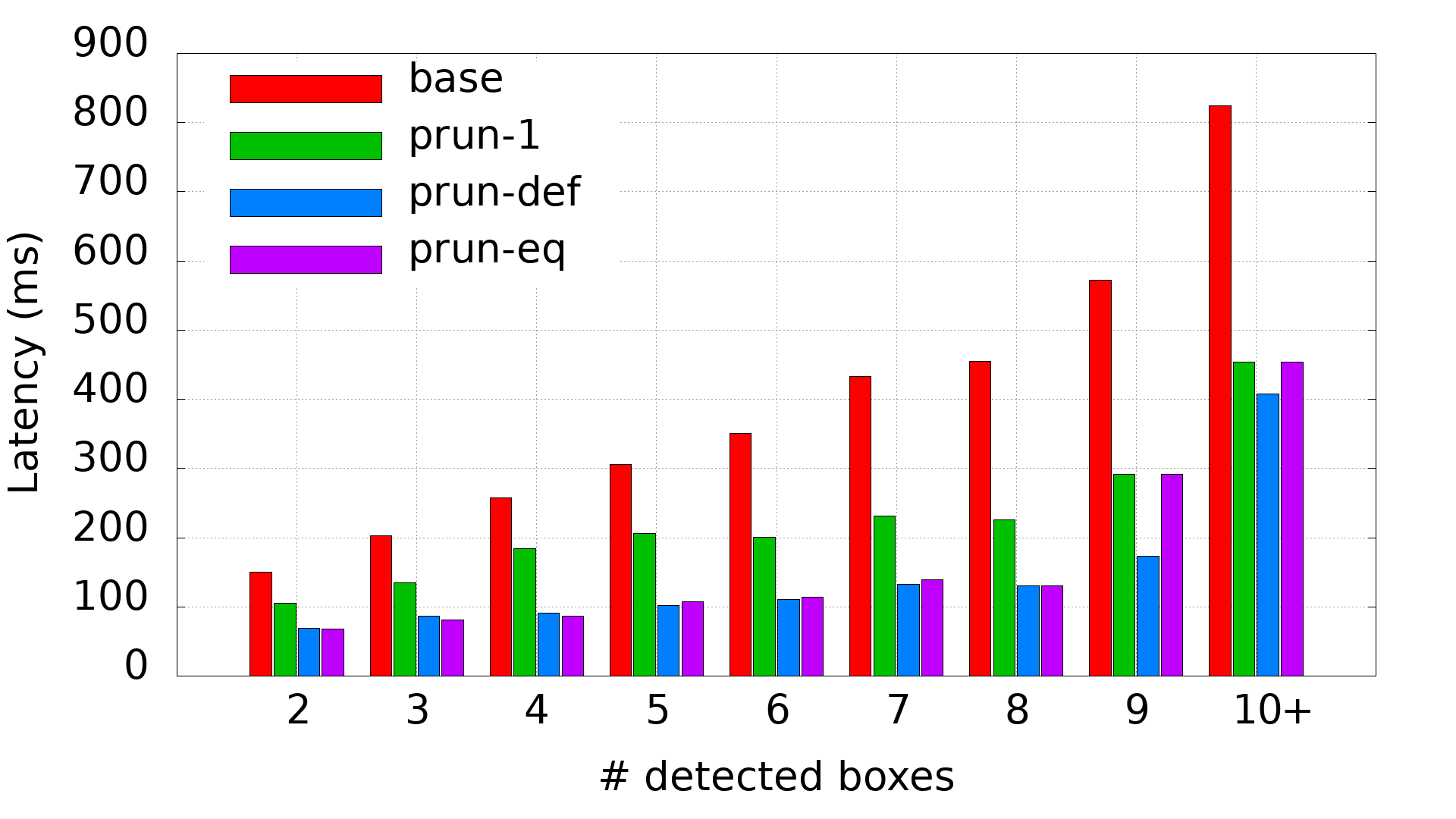}}
\subfloat[][End-to-End Inference]{\includegraphics[width=0.33\linewidth]{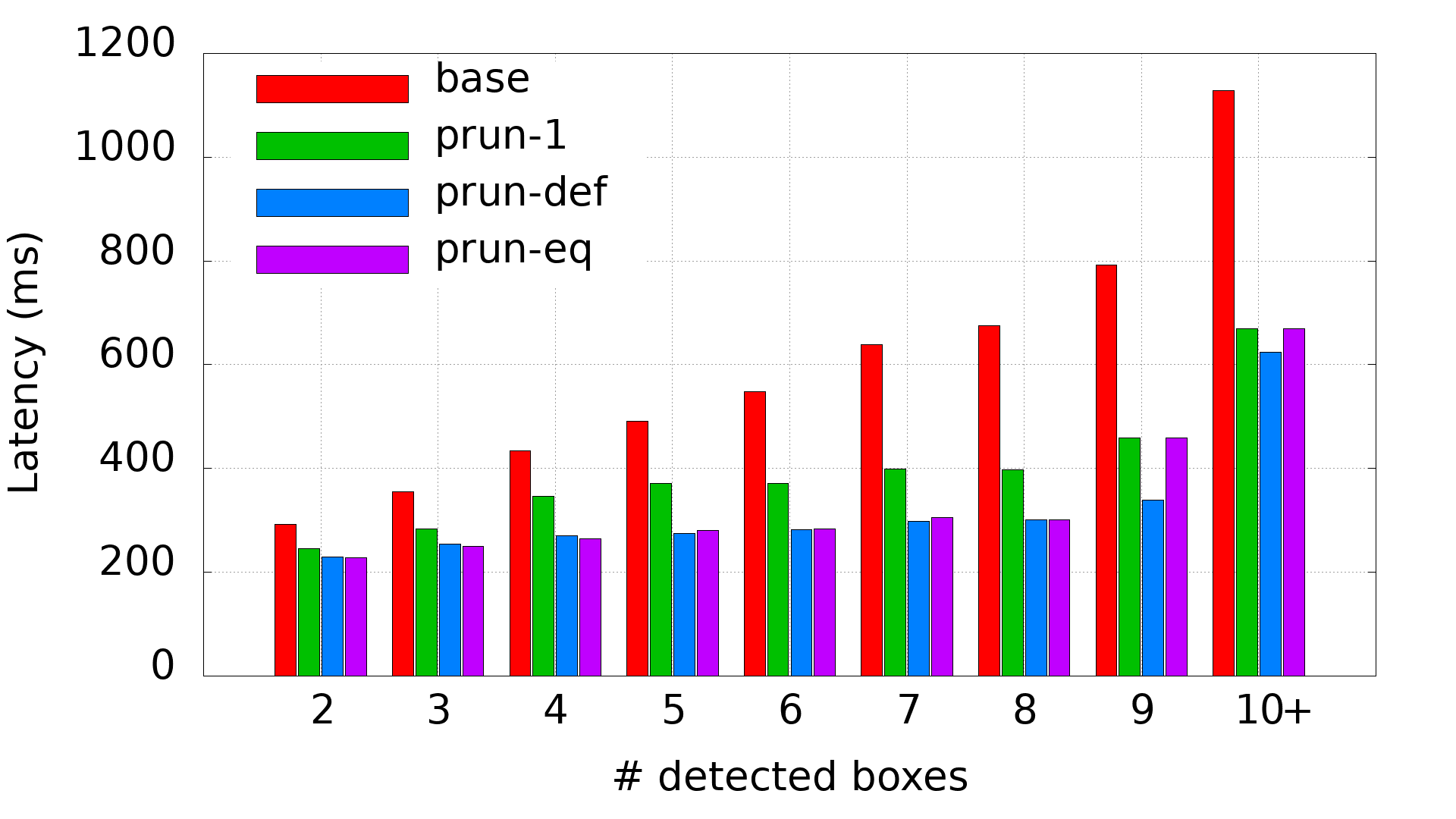}}
\caption{The impact of using \code{prun} in PaddleOCR.}
\label{fig:paddleOCR-prun-cmp}
\end{figure*}

In light of the discussion above, we break down the comparison of the latency results 
by the number of detected boxes, as depicted in Figure~\ref{fig:paddleOCR-prun-cmp}. 
The latency numbers in this figure were collected with $16$ cores; we discuss the overall scalability trends later on.
We also break down the performance in two of the phases where we
have used \code{prun}, namely Text Classification (Figure~\ref{fig:paddleOCR-prun-cmp}~(a)) and Recognition (Figure~\ref{fig:paddleOCR-prun-cmp}~(b)).

Considering the results in Figure~\ref{fig:paddleOCR-prun-cmp}, 
one can notice that, as expected, the benefit of \code{prun} increases with the number of detected text boxes.
For instance, when considering the total end-to-end latency (Figure~\ref{fig:paddleOCR-prun-cmp}~(c)),
with only two boxes \code{prun-def} outperforms \code{base} by $1.28$x.
However, with $9$ and $10+$ boxes,  \code{prun-def} outperforms \code{base} by $2.33$x and $1.81$x, respectively.

It is interesting to compare the performance of \code{prun-def} with other \code{pun}-based variants.
As one can notice in Figure~\ref{fig:paddleOCR-prun-cmp}~(a), the \code{prun-1} variant produces the lowest latency
when the number of detected boxes is small.
In fact, the \code{base} variant also performs better than \code{prun-def} in this case.
We attribute this to two factors.
First, this specific phase of the pipeline shows negative scalability, 
which can be also seen in Figure~\ref{fig:paddleOCR-base}.
Therefore, best performance is achieved when fewer threads per box is used in this phase, 
which is what \code{prun-1} effectively achieves.
Second, \code{prun-def} (and \code{prun-eq}) create and destroy more threads than \code{prun-1} 
in those cases as they create thread pools containing more threads for each \code{prun} invocation.
This adds small, but non-negligible overhead given that the the execution time of this phase is short.
In the future work, we intend to experiment with reusing thread pools between \code{prun} invocations.
As the number of detected boxes increases, however, all \code{prun} variants allocate less threads (or even just $1$) per each box, 
and they allocate a similar number of threads for their pools, thus closing the gap with the \code{prun-1} variant.

\begin{figure}[t!]
\includegraphics[width=1\linewidth]{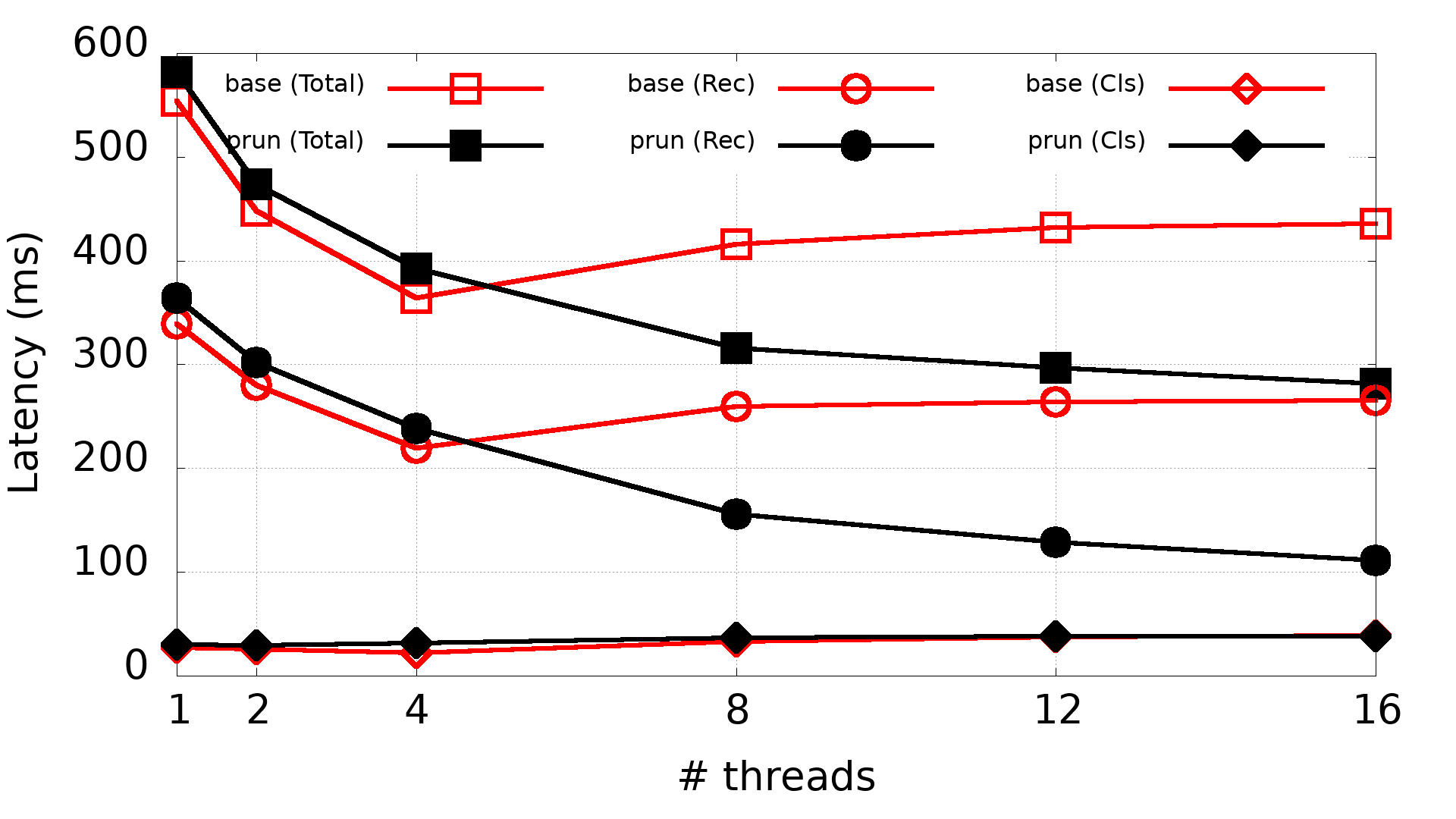}
\caption{Total (end-to-end) inference latency of PaddleOCR with a varying number of threads. Also shown the latency of Text Classification (Cls) and Text Recognition (Rec) phases}
\label{fig:paddleOCR-all}
\end{figure}

When the Text Recognition phase is concerned (cf.~Figure~\ref{fig:paddleOCR-prun-cmp}~(b)), however, it is apparent 
from Figure~\ref{fig:paddleOCR-base} that one can improve its latency by using more than one thread.
We note that, quantitively, this phase is also far more dominant than the Text Classification one.
Here, \code{prun-def} manages to achieve best or close to best result across all counts of detected boxes, which
translates to overall highly competitive end-to-end inference performance (cf.~Figure~\ref{fig:paddleOCR-prun-cmp}~(c)).
In general, the results in Figure~\ref{fig:paddleOCR-prun-cmp} call for a dynamic mechanism, which would choose the best thread allocation strategy based on the
given workload and available resources.
Devising and experimenting with such a strategy is left for future work.

Finally, we shed more light on how the scalability improves with the use of \code{prun} in Figure~\ref{fig:paddleOCR-all}, 
where we vary the number of cores (and therefore, the total number of worker threads) available for OnnxRuntime.
Once again, we include the latency of each of the two last phases of PaddleOCR (denoted as Rec for Text Recognition and Cls for Text Classification) 
along with the end-to-end (Total) latency.
We include only the results of the \code{base} and \code{prun-def} variants (denoted simply as \code{prun} in Figure~\ref{fig:paddleOCR-all}), for clarity.

Overall, one can notice similar trends to the ones discussed above.
In the \code{base} version, the Text Recognition phase does scale up to $4$ threads, but then its performance suffers as the number
of threads increases.
The \code{prun} variant avoids this performance degradation, and in fact, continues to scale up to $16$ threads.
Indeed, when considering the Text Recognition phase only, the \code{prun} variant outperforms \code{base} by more than $2.4$x at $16$ threads.
However, since both variants have an identical Text Detection phase, which according to Figure~\ref{fig:paddleOCR-base} subsumes a 
substantial part of the total latency, the end-to-end speedup of \code{prun} is only $1.5$x at $16$ threads.

\subsection{Batching of Heterogeneous Inputs}
\label{sec:hetero-batch}
Our next example concerns with the Transformer architecture~\cite{VSP17}, which 
revolutionized the domain of NLP when it was introduced in 2017 and 
has been applied to other domains since then (e.g.,~\cite{DBK21, LPW20}).
This architecture consists of a stack of layers, each composed of 
a self-attention block followed by a fully connected network~\cite{VSP17}.
Past work has shown that the majority of computation cycles in Transformers is
spent on (scalable) matrix multiplication operations, yet up to one third of 
the cycles is spent elsewhere (i.e., less scalable operations)~\cite{DK21}.

It is well-known that one way to improve the inference performance (specifically, throughput) 
of Transformers is through \emph{input batching}~\cite{WWW21, ARZ22, FYZ21}.
This strategy works well, however, when the inputs have the same length.
Otherwise, one has either give up on batching, or pad inputs to the same length.
The latter results in wasted computation cycles, since special padding tokens are treated
exactly as input tokens by the architecture and dismissed at the end of the computation.

This situation presents an ideal case for applying the Divide-and-Conquer Principle.
Instead of padding the inputs of various lengths up to the longest input in the batch, 
we can run inference on those inputs (as they are, without padding) using the \code{prun} API, and let the runtime decide
how many cores should be used to process each of the inputs.
We modify the Transformer benchmark built into the OnnxRuntime~\cite{OnnxRuntime-transformers-bench} to implement this strategy.

To evaluate the effectiveness of the approach described above, we set up an experiment
where we generate $X$ inputs of a length chosen uniformly and randomly out of the range $[16, 512]$.
We then compare the \code{pad-batch} version in which all $X$ inputs are padded to the 
longest length in the given batch with the \code{prun} version in which the inference is invoked 
with \code{prun} on all inputs in the batch.
We show results with the highly popular BERT model~\cite{DCL19} (specifically, \code{``bert-based-uncased''}).
We have also experimented with other Transformer-based models (such as ``bert-large-uncased'' or ``roberta-base'')
measuring similar qualitative results.

\begin{figure}[t!]
\includegraphics[width=1\linewidth]{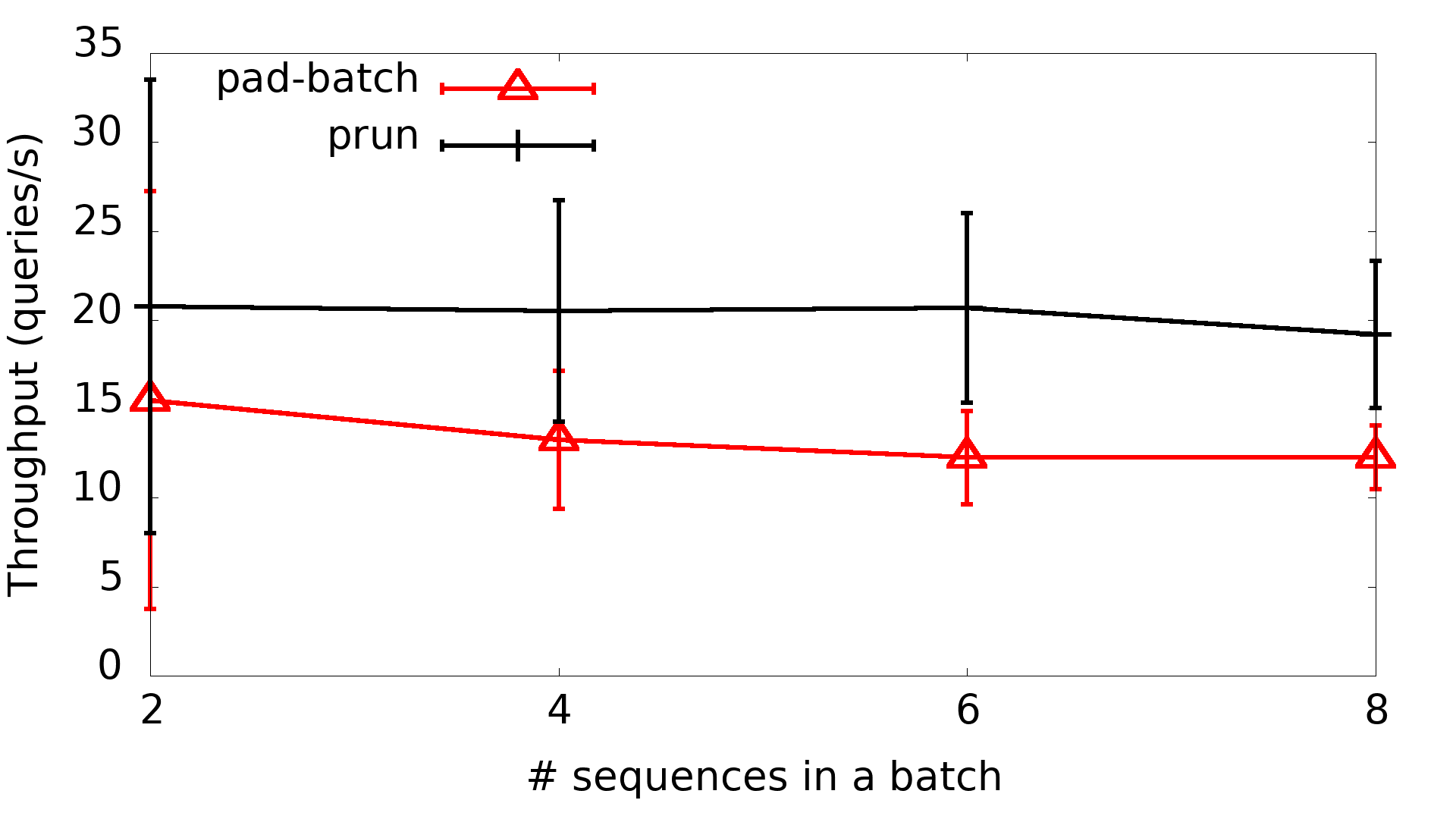}
\caption{Throughput of inferencing BERT on batches of sequences of sizes chosen randomly from the range $[16, 512]$}
\label{fig:transformers-random-batches}
\end{figure}

We note that this experiment includes inherent amount of randomness --- a batch of small sentences is as likely to be chosen as a 
batch of long sentences.
In an attempt to reduce the anticpated high varaince of the results, 
we opted to repeat the experiment $1000$ times, and so for each $X$, each data point is an average of $1000$ results.
Figure~\ref{fig:transformers-random-batches} presents the throughput results with batches of various sizes (i.e., $X$ varies from $2$ to $8$),
with error bars depicting the standard deviation of the reported mean.
Even though \code{prun} outperforms the \code{pad-batch} variant across all batch sizes, the variance in 
the measured results remains exceptionally high.

\begin{figure}[t!]
\includegraphics[width=1\linewidth]{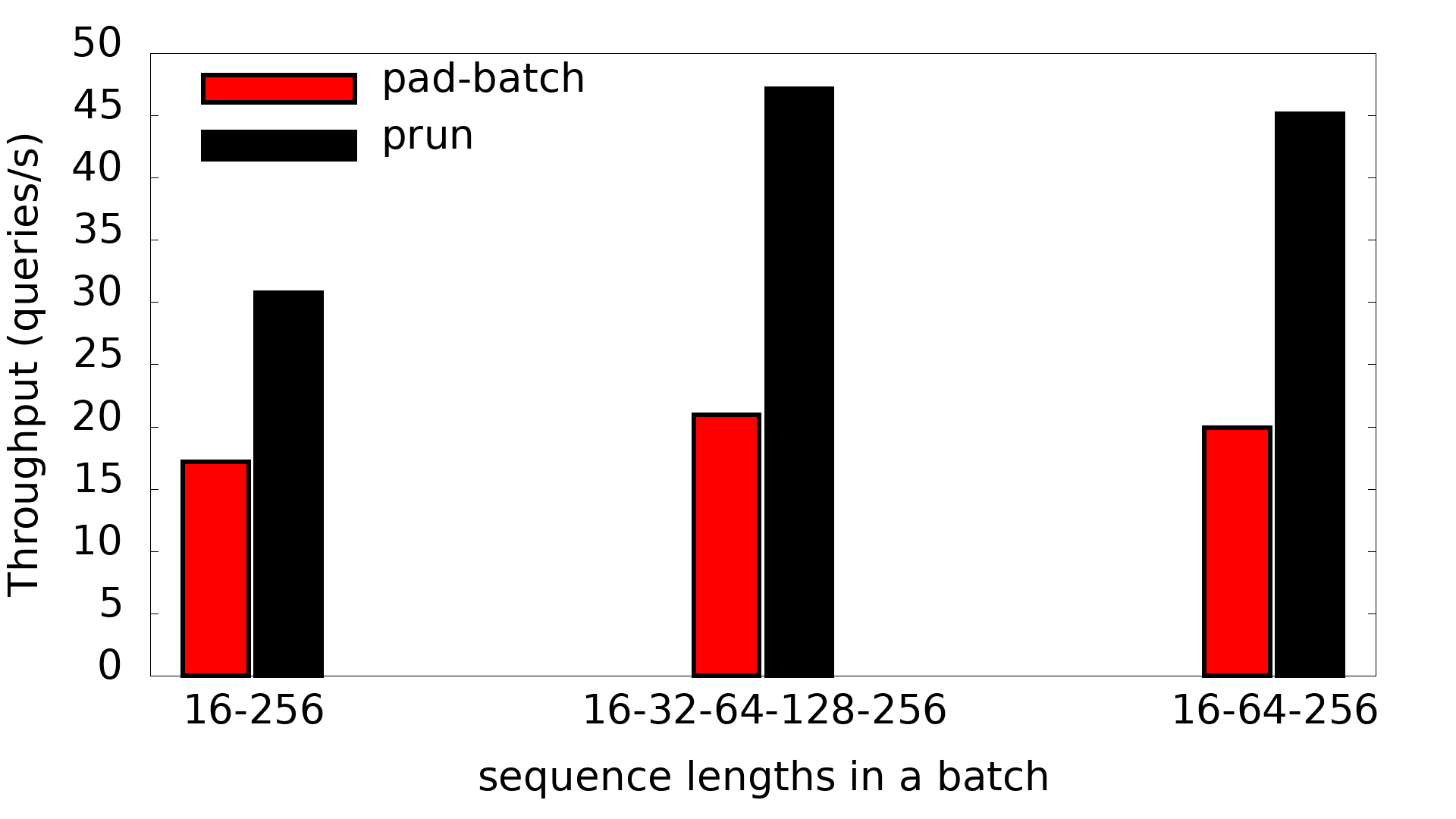}
\caption{Throughput of inferencing BERT on batches of sequences of various preset sizes}
\label{fig:transformers-preset-sizes}
\end{figure}

As a result, we setup two additional experiments in a more controlled way likely to produce more stable results.
In the first, we simply preset the lengths of various sequences in each batch.
For instance, a batch denoted as ``$16$-$64$-$256$'' includes three sentences, one is $16$, another is $64$ and yet another is $256$ tokens long.
We show the results of this experiment in Figure~\ref{fig:transformers-preset-sizes}.
Here, the \code{prun} version easily outperforms the \code{pad-batch} variant, which has to pad all sequences to the longest sequence in a batch.
As one might expect, the benefit from using \code{prun} increases with the number of sentences in a batch, as 
this variant eliminates all the redundant work associated with padding.

In the second experiment, we use a batch of $1$ long sentence ($256$ tokens long) and $X$ short sequences of $16$ tokens each,
where we vary $X$ between $0$ and $15$.
We show the throughput results of this experiment in Figure~\ref{fig:transformers-preset-sizes}, along with a curve
depicting the number of threads allocated by \code{prun} for the long sequence in the batch.

There are several interesting observations that can be made here. 
First, when $X$=0, i.e., the batch contains only one long sentence, 
both variants employ all available cores to process that batch, producing similar result.
This shows that the overhead of using \code{prun} when the input has only one chunk is negligible.
Second, the throughput of the \code{pad-batch} version grows, but modestly with the increase in the number of short sequences.
This is because, as stated above, a larger batch of (padded) sequences helps to achieve better throughput with Transformers.
At the same time, the throughput growth with \code{prun} is much more dramatic up to $3$ short sequences in a batch and then
it declines, but stays well above that achieved with \code{pad-batch}.
Both phenomena can be explained with the fact that inferencing a sequence of $256$ tokens takes 
about the same time with $16$ threads as it takes with $13$.
Thus, adding a few short sequences into the batch, each allocated with just $1$ thread (as they have small relative weight),
has negligible impact on the latency, but improves throughput.
With more short sequences in a batch, less threads are allocated for the long sequence 
(as can be seen in Figure~\ref{fig:transformers-preset-sizes}) and its inference latency grows.
This causes the overall throughput to decrease.

\begin{figure}[t!]
\includegraphics[width=1\linewidth]{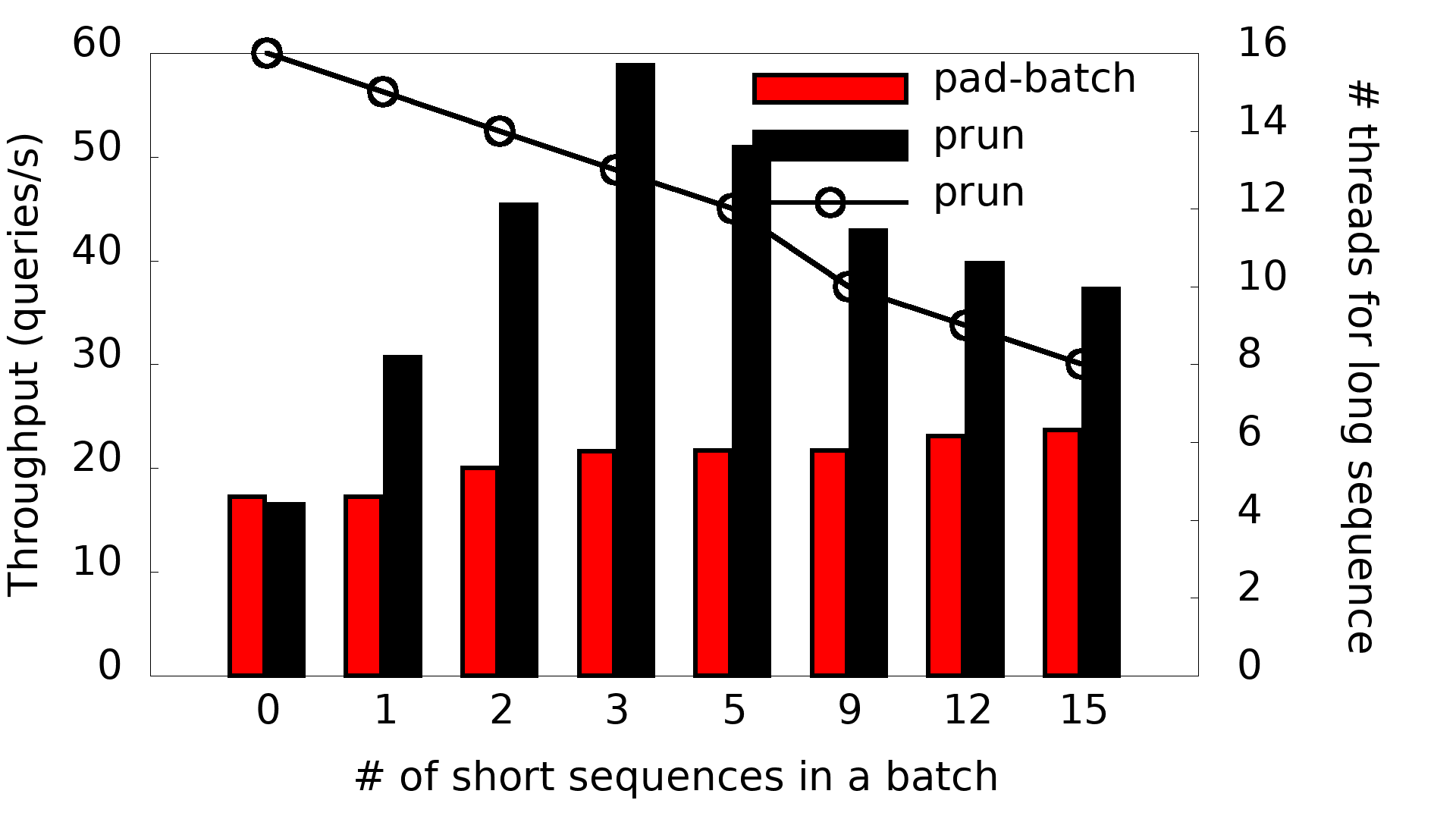}
\caption{Throughput of inferencing BERT on a batch containing one long sentence of $256$ tokens and $X$ short sequence with $16$ tokens each, where $X$ varies $0$ to $15$. In addition, we show how many threads are dedicated to the inference of the one long sentence in the batch in the \code{prun} variant}
\label{fig:transformers-two-sizes}
\end{figure}

\subsection{Batching of Homogeneous Inputs}
Our last example follows directly from the discussion in Section~\ref{sec:why-slow} on the lack of scalability in ML models.
As already mentioned, while Transformers models heavily use scalable matrix multiplication 
operations, they also employ less scalable operations. 
The impact of the latter grows with the increase in the number of cores.
Therefore, one may benefit form the Divide-and-Conquer Principle applied to Transformers 
\emph{even when the batch includes inputs of the same length}.

As a concrete example, consider a batch of two inputs.
Instead of using all available cores to process the batch,
we will use half the cores for each input.
Intuitively, the less scalable operators create less relative 
overhead when less cores are used and the input sequence is shorter 
(i.e., contains half the tokens compared to the entire batch).

\begin{figure}[t!]
\includegraphics[width=1\linewidth]{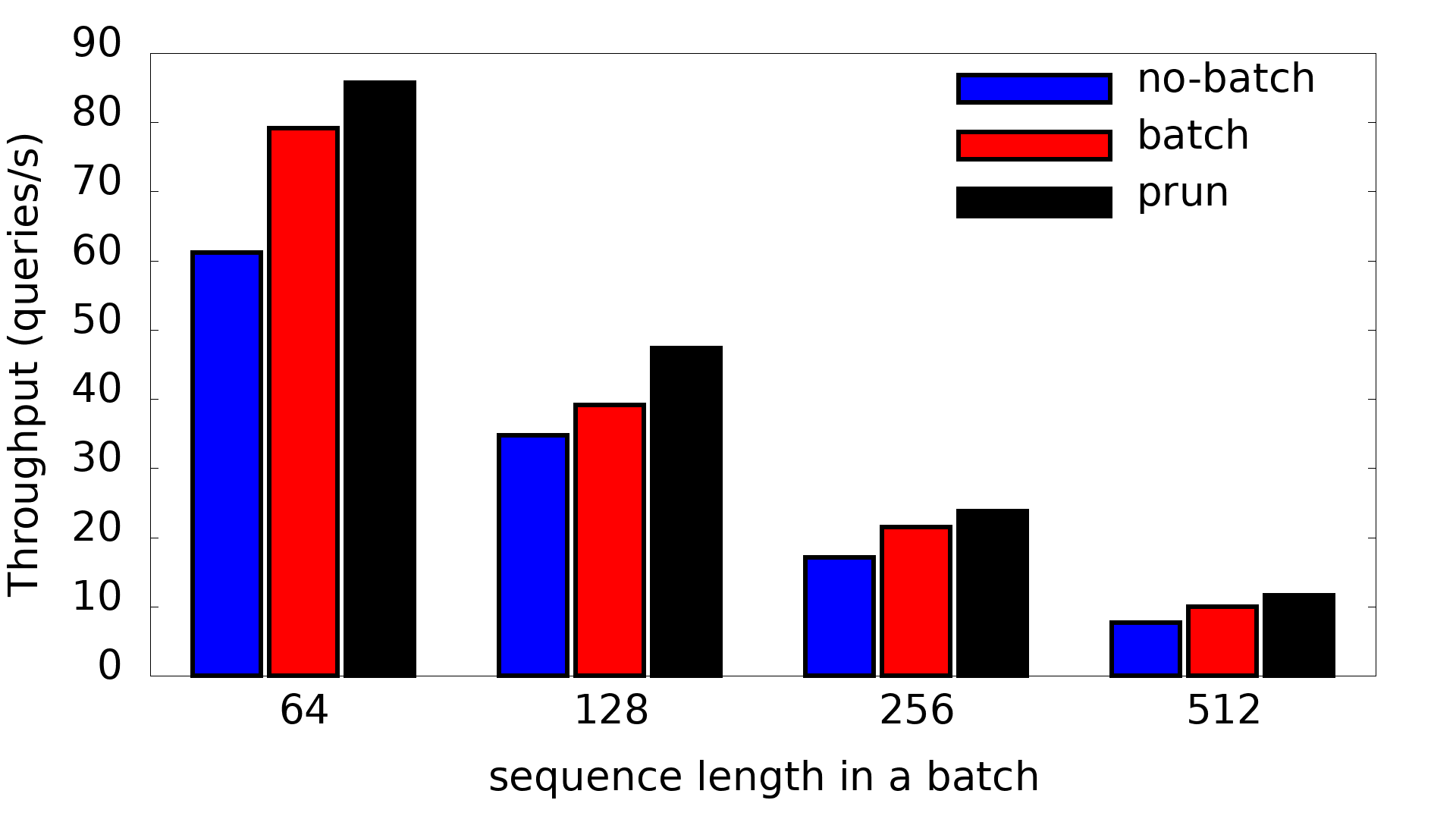}
\caption{Throughput of inferencing BERT with batches of $4$ sequences of equal size}
\label{fig:transformers-equal-batches}
\end{figure}

Figure~\ref{fig:transformers-equal-batches} demonstrates this effect with batches of inputs of equal lengths.
In addition to the \code{pad-batch} variant (which we simply call \code{batch} here, as no
padding is required) and \code{prun}, we include a \code{no-batch} variant, which runs inference
on each sequence in a given batch one at a time.
Note that we include the latter to simply demonstrate the benefits of batching in general,
confirming previous findings~\cite{WWW21, ARZ22, FYZ21}.
Each set of bars in Figure~\ref{fig:transformers-equal-batches} corresponds to a batch of $4$ sentences with the
given length (from $64$ tokens to $512$).
Overall, the \code{prun} version yields a more modest (yet non-trivial) speedup over \code{batch} compared to the
case of non-homogeneous inputs in Section~\ref{sec:hetero-batch}.
This is expected, since in this case the room for improvement (over \code{batch}) does not include wasted
computation related to padding.

\section{Related Work}
\label{sec:related}

As mentioned in the Introduction, the major focus of the ML community has been on 
improving the accuracy and training performance of proposed models, while
efficient inferencing and serving of those models receives relatively less attention. 
Yet, there have been some notable exceptions of work focused specifically on inference performance, and
we survey the most relevant results hereafter.
As an aside, we note that many of the results below come from less formal blog posts published by various companies,
highlighting the great practical importance of efficient inference.

Wang et al.~\cite{WWW21} explore various factors that influence inference performance in TensorFlow, 
including the choice of a specific math library, a thread pool library, availability of SIMD (single instruction multiple data) support, etc. 
They identify data preparation as one of the causes for poor scalability of small matrix multiplication operations,
something we more generally attribute to framework overhead in Section~\ref{sec:why-slow}.
They come up with a set of guidelines one can use to tune TensorFlow settings to achieve better performance
compared to the one achieved with settings recommended by TensorFlow authors or Intel.

With the tremendous rise in popularity of Transformers, 
several papers and blog posts focus on its inference performance.
Dice and Kogan investigate inference performance of Transformers on CPUs~\cite{DK21}.
Their analysis shows that most inference computation cycles are spent in matrix multiplication 
operations. 
Hence, they propose an adaptive matrix multiplication optimization aimed at reducing
the latency of those operations and subsequently improving the overall inference performance.
Intel engineers describe an effort to optimize inference of BERT in Apache MXNet using the GluonNLP toolkit,
where one of the ideas is to quantize the model for better performance with lower precision~\cite{NYZ19}.
Similar quantization ideas (along with \emph{distillation}, another common method of 
reducing the size of a model~\cite{SDC19}) were employed by Roblox to speedup their deployment of BERT on CPUs~\cite{LK20}.
The same blog post also mentions that eliminating padding of input sentences has led to better performance (though the
authors did that for batches of $1$ input only).
A Microsoft team~\cite{NYZ20} describes their effort on accelerating BERT with OnnxRuntime
through operation fusion that helps to reduce the amount of overhead (e.g., memory copying) in invoking
each kernel individually.

A few recent papers and projects have looked into the deficiency of padding of heterogenous inputs.
Fang et al.~\cite{FYZ21} propose a sequence-length-aware batch scheduler, which aims to
batch requests of a similar size, thus reducing the cost of zero padding of all requests into one batch.
It requires a profiling phase during which the inference cost of various batches is collected.
Du et al.~\cite{DJU22} propose to carefully redesign the GPU kernels employed by Transformers
to eliminate most redundant computation associated with zero padding.
The Effective Transformer project by ByteDance~\cite{EffectiveTransformer} aims to dynamically remove and restore padding
during different calculation stages.
All those efforts target specifically the inferencing Transformers on GPUs, and
it is not clear how efficient they would be on CPUs and/or with other architectures.

Beyond Transformers, Liu at et.~\cite{LWY19} describe NeoCPU, an approach for optimizing CNN inference on CPUs. 
NeoCPU proposes a configurable design of an efficient convolution operation that can be tuned efficiently to popular CPUs.
This design is coupled with a scheme for obtaining the best memory layout for data in different operations of a CNN model,
in order to minimize the overhead of transforming the data between various individual operations.

\section{Discussion}
\label{sec:discussion}
In this paper, we have discussed various reasons for the lack of scalability of inferencing ML models.
While the reasons vary from micro to macro-levels, the common motive is that existing ML
frameworks are geared towards high performance training.
This is expressed by the fact that kernels for common operations are typically optimized for 
large batches with long inputs, ignoring relatively small overheads in various parts of those frameworks 
that are immaterial to the overall training performance.
However, during inference the batches tend to be much smaller and contain shorter inputs,
thus making those overheads more prominent.
A somewhat similar observation has been made by Aminabadi et al.~\cite{ARZ22}.

We leverage this poor scalability and describe a simple, yet powerful approach, in which
the given input is broken into chunks and each chunk is processed in parallel, instead of using
all available resources for the entire input.
As we demonstrate with a few well-known models, this approach improves inference scalability
and ultimately can lead to over $2$x latency and throughput improvements.

This work offers several directions for future research.
First, we want to explore more dynamic thread allocation strategies, e.g., 
ones that can better adjust to the cases where the weight of a work chunk does
not correlate linearly with its size and/or where the underlying model performs best
while running with a single thread.
Second, we want to find ways to automate splitting the input into chunks that can be
processed in parallel, lowering the cost (in terms of user code changes) of using \code{prun}
even further.
Finally, we want to explore other use cases where the use of \code{prun} would be beneficial,
including other ML models that feature a pipeline-based architecture (e.g.,~\cite{WS15, MHL21}).

\begin{acks}
The author would like to thank Dave Dice for valuable comments on an early draft of this paper.
\end{acks}

\remove{
Outline
1. Inference of Machine Learning models is not scalable (example).
2. Reasons and possible solutions
2.1 "Not enough work"
2.2. Sequential operators
2.3. ML Framework overhead
3. What we do in this paper
4. Additional use-cases
4.1. batching of queries with different lengths
}

 
\bibliographystyle{plain}
\bibliography{scale-inference}

\begin{thebibliography}{10}

\bibitem{APY20}
Ahsan Ali, Riccardo Pinciroli, Feng Yan, and Evgenia Smirni.
\newblock Batch: Machine learning inference serving on serverless platforms
  with adaptive batching.
\newblock In {\em Proceedings of the International Conference for High
  Performance Computing, Networking, Storage and Analysis (SC)}, 2020.

\bibitem{Amd67}
Gene~M. Amdahl.
\newblock Validity of the single processor approach to achieving large scale
  computing capabilities.
\newblock In {\em Proceedings of the Spring Joint Computer Conference (AFIPS)},
  page 483–485, 1967.

\bibitem{ARZ22}
Reza~Yazdani Aminabadi, Samyam Rajbhandari, Minjia Zhang, Ammar~Ahmad Awan,
  Cheng Li, Du~Li, Elton Zheng, Jeff Rasley, Shaden Smith, Olatunji Ruwase, and
  Yuxiong He.
\newblock {DeepSpeed} inference: Enabling efficient inference of transformer
  models at unprecedented scale.
\newblock {\em CoRR}, abs/2207.00032, 2022.

\bibitem{BKH16}
Jimmy~Lei Ba, Jamie~Ryan Kiros, and Geoffrey~E. Hinton.
\newblock Layer normalization.
\newblock {\em CoRR}, abs/1607.06450, 2016.

\bibitem{BMR20}
Tom~B. Brown, Benjamin Mann, Nick Ryder, Melanie Subbiah, Jared Kaplan,
  Prafulla Dhariwal, Arvind Neelakantan, Pranav Shyam, Girish Sastry, Amanda
  Askell, Sandhini Agarwal, Ariel Herbert{-}Voss, Gretchen Krueger, Tom
  Henighan, Rewon Child, Aditya Ramesh, Daniel~M. Ziegler, Jeffrey Wu, Clemens
  Winter, Christopher Hesse, Mark Chen, Eric Sigler, Mateusz Litwin, Scott
  Gray, Benjamin Chess, Jack Clark, Christopher Berner, Sam McCandlish, Alec
  Radford, Ilya Sutskever, and Dario Amodei.
\newblock Language models are few-shot learners.
\newblock In {\em Advances in Neural Information Processing Systems (NeurIPS)},
  2020.

\bibitem{EffectiveTransformer}
ByteDance.
\newblock {Effective Transformer}.
\newblock \url{https://github.com/bytedance/effective\_transformer}.
\newblock Accessed: 07-29-22.

\bibitem{TorchServe}
PyTorch~Serve Contributors.
\newblock {TorchServe}.
\newblock \url{https://pytorch.org/serve}.
\newblock Accessed: 07-28-22.

\bibitem{CLR09}
Thomas~H. Cormen, Charles~E. Leiserson, Ronald~L. Rivest, and Clifford Stein.
\newblock {\em Introduction to Algorithms, 3rd Edition}.
\newblock {MIT} Press, 2009.

\bibitem{CWZ17}
Daniel Crankshaw, Xin Wang, Giulio Zhou, Michael~J. Franklin, Joseph~E.
  Gonzalez, and Ion Stoica.
\newblock Clipper: {A} low-latency online prediction serving system.
\newblock In {\em {USENIX} Symposium on Networked Systems Design and
  Implementation ({NSDI})}, pages 613--627, 2017.

\bibitem{DCL19}
Jacob Devlin, Ming{-}Wei Chang, Kenton Lee, and Kristina Toutanova.
\newblock {BERT:} pre-training of deep bidirectional transformers for language
  understanding.
\newblock In {\em Proc. of NAACL-HLT}, pages 4171--4186, 2019.

\bibitem{DK21}
Dave Dice and Alex Kogan.
\newblock Optimizing inference performance of {Transformers} on {CPUs}.
\newblock In {\em Workshop on Machine Learning and Systems (EuroMLSys)}, 2021.

\bibitem{DBK21}
Alexey Dosovitskiy, Lucas Beyer, Alexander Kolesnikov, Dirk Weissenborn,
  Xiaohua Zhai, Thomas Unterthiner, Mostafa Dehghani, Matthias Minderer, Georg
  Heigold, Sylvain Gelly, Jakob Uszkoreit, and Neil Houlsby.
\newblock An image is worth 16x16 words: Transformers for image recognition at
  scale.
\newblock {\em ICLR}, 2021.

\bibitem{DJU22}
Jiangsu Du, Jiazhi Jiang, Yang You, Dan Huang, and Yutong Lu.
\newblock Handling heavy-tailed input of transformer inference on gpus.
\newblock In {\em Proc. of ACM International Conference on Supercomputing
  (ICS)}, 2022.

\bibitem{DLG20}
Yuning Du, Chenxia Li, Ruoyu Guo, Xiaoting Yin, Weiwei Liu, Jun Zhou, Yifan
  Bai, Zilin Yu, Yehua Yang, Qingqing Dang, and Haoshuang Wang.
\newblock {PP-OCR:} {A} practical ultra lightweight {OCR} system.
\newblock {\em CoRR}, abs/2009.09941, 2020.

\bibitem{FYZ21}
Jiarui Fang, Yang Yu, Chengduo Zhao, and Jie Zhou.
\newblock {TurboTransformers}: an efficient {GPU} serving system for
  transformer models.
\newblock In {\em Proc. of {ACM SIGPLAN PPoPP}}, pages 389--402, 2021.

\bibitem{TFServing}
Google.
\newblock {Tensorflow Serving}.
\newblock \url{https://www.tensorflow.org/tfx/guide/serving}.
\newblock Accessed: 07-28-22.

\bibitem{OpenImages}
Ivan Krasin, Tom Duerig, Neil Alldrin, Vittorio Ferrari, Sami Abu-El-Haija,
  Alina Kuznetsova, Hassan Rom, Jasper Uijlings, Stefan Popov, Shahab Kamali,
  Matteo Malloci, Jordi Pont-Tuset, Andreas Veit, Serge Belongie, Victor Gomes,
  Abhinav Gupta, Chen Sun, Gal Chechik, David Cai, Zheyun Feng, Dhyanesh
  Narayanan, and Kevin Murphy.
\newblock {OpenImages}: A public dataset for large-scale multi-label and
  multi-class image classification.
\newblock {\em Dataset available from
  https://storage.googleapis.com/openimages/web/index.html}, 2017.

\bibitem{LPW20}
Hang Le, Juan~Miguel Pino, Changhan Wang, Jiatao Gu, Didier Schwab, and Laurent
  Besacier.
\newblock Dual-decoder transformer for joint automatic speech recognition and
  multilingual speech translation.
\newblock In {\em Proceedings of the International Conference on Computational
  Linguistics ({COLING})}, pages 3520--3533, 2020.

\bibitem{LK20}
Quoc~N. Le and Kip Kaehler.
\newblock {How We Scaled Bert To Serve 1+ Billion Daily Requests on CPUs}.
\newblock
  \url{https://blog.roblox.com/2020/05/scaled-bert-serve-1-billion-daily-requests-cpus}.
\newblock Published: 05-27-20, Accessed: 08-02-22.

\bibitem{LWY19}
Yizhi Liu, Yao Wang, Ruofei Yu, Mu~Li, Vin Sharma, and Yida Wang.
\newblock Optimizing {CNN} model inference on cpus.
\newblock In {\em Proc. of {USENIX} Annual Technical Conference (ATC)}, pages
  1025--1040, 2019.

\bibitem{MHL21}
Matteo Maggioni, Yibin Huang, Cheng Li, Shuai Xiao, Zhongqian Fu, and Fenglong
  Song.
\newblock Efficient multi-stage video denoising with recurrent spatio-temporal
  fusion.
\newblock In {\em {IEEE} Conference on Computer Vision and Pattern Recognition,
  {CVPR}}, pages 3466--3475, 2021.

\bibitem{MVG07}
Bryan Marker, Field~G. Van~Zee, Kazushige Goto, Gregorio Quintana-Ort\'{\i},
  and Robert~A. van~de Geijn.
\newblock Toward scalable matrix multiply on multithreaded architectures.
\newblock In {\em Proceedings of the International Conference on Parallel
  Processing (EuroPar)}, page 748–757, 2007.

\bibitem{MAH16}
Ian Masliah, Ahmad Abdelfattah, A.~Haidar, S.~Tomov, Marc Baboulin, J.~Falcou,
  and J.~Dongarra.
\newblock High-performance matrix-matrix multiplications of very small
  matrices.
\newblock In {\em Proceedings of the International Conference on Parallel
  Processing (EuroPar)}, page 659–671, 2016.

\bibitem{OnnxRuntime}
Microsoft.
\newblock {OnnxRuntime}.
\newblock \url{https://onnxruntime.ai}.
\newblock Accessed: 08-02-22.

\bibitem{OnnxRuntime-transformers-bench}
Microsoft.
\newblock {Transformer Model Optimization Tool Overview}.
\newblock
  \url{https://github.com/microsoft/onnxruntime/tree/master/onnxruntime/python/tools/transformers}.
\newblock Accessed: 08-02-22.

\bibitem{NYZ20}
Emma Ning, Nathan Yan, Jeffrey Zhu, and Jason Li.
\newblock Microsoft open sources breakthrough optimizations for transformer
  inference on gpu and cpu.
\newblock
  \url{https://cloudblogs.microsoft.com/opensource/2020/01/21/microsoft-onnx-open-source-optimizations-transformer-inference-gpu-cpu/}.
\newblock Published: 01-20-20, Accessed: 01-06-21.

\bibitem{SDC19}
Victor Sanh, Lysandre Debut, Julien Chaumond, and Thomas Wolf.
\newblock Distilbert, a distilled version of {BERT:} smaller, faster, cheaper
  and lighter.
\newblock {\em CoRR}, abs/1910.01108, 2019.

\bibitem{VSP17}
Ashish Vaswani, Noam Shazeer, Niki Parmar, Jakob Uszkoreit, Llion Jones,
  Aidan~N. Gomez, Lukasz Kaiser, and Illia Polosukhin.
\newblock Attention is all you need.
\newblock In {\em Proc. of NeurIPS}, pages 5998--6008, 2017.

\bibitem{WS15}
Li~Wang and Dennis Sng.
\newblock Deep learning algorithms with applications to video analytics for a
  smart city: a survey.
\newblock {\em CoRR}, abs/1512.03131, 2015.

\bibitem{WWW21}
Yu~Emma Wang, Carole-Jean Wu, Xiaodong Wang, Kim Hazelwood, and David Brooks.
\newblock Exploiting parallelism opportunities with deep learning frameworks.
\newblock {\em ACM Trans. Archit. Code Optim.}, 18(1), 2021.

\bibitem{NYZ19}
Shufan Wu, Tao Lv, Pengxin Yuan, Patric Zhao, Jason Ye, and Haibin Lin.
\newblock {Optimization for BERT Inference Performance on CPU}.
\newblock
  \url{https://medium.com/apache-mxnet/optimization-for-bert-inference-performance-on-cpu-3bb2413d376c}.
\newblock Published: 09-12-19, Accessed: 08-02-22.

\bibitem{ZWZ22}
Zhe Zhou, Xuechao Wei, Jiejing Zhang, and Guangyu Sun.
\newblock {PetS: A Unified Framework for Parameter-Efficient Transformers
  Serving}.
\newblock In {\em {USENIX} Annual Technical Conference (ATC)}, 2022.

\end{thebibliography}

\end{document}